\documentclass[1p]{elsarticle}

\usepackage{hyperref}
\usepackage{amsmath,epsfig}
\usepackage{graphicx}
\usepackage{epstopdf}
\usepackage{booktabs}
\usepackage{caption}
\usepackage{float}
\usepackage{amsmath}
\usepackage{algorithm}
\usepackage{algorithmic}
\usepackage{subfigure}
\usepackage{nth}
\usepackage{amssymb}

\journal{Journal of Neurocomputing}

\DeclareMathOperator*{\argmin}{\argmin}

\newcommand{\INPUT}{\item[\myinput]}
\newcommand{\myinput}{\textbf{Initialization:}}

\newcommand{\MYWHILE}{\item[\mywhile]}
\newcommand{\mywhile}{\textbf{Repeat}}

\newcommand{\MYENDWHILE}{\item[\myendwhile]}
\newcommand{\myendwhile}{\textbf{until}}

\begin{document}

\begin{frontmatter}

\title{Deep Boosting: Joint Feature Selection and Analysis Dictionary Learning in Hierarchy\tnoteref{t1}}

\author[mymainaddress]{Zhanglin Peng}
\author[mymainaddress]{Ya Li}
\author[mysecondaryaddress]{Zhaoquan Cai}
\author[mymainaddress]{Liang Lin\corref{cor1}}

\address[mymainaddress]{Sun Yat-sen University, Guangzhou, China}
\address[mysecondaryaddress]{Huizhou University, Huizhou, China}

\cortext[cor1]{Corresponding author: Liang Lin (linliang@ieee.org).}
\tnotetext[t1]{Project page: \url{http://vision.sysu.edu.cn/projects/deepboosting/}}

\begin{abstract}
This work investigates how the traditional image classification pipelines can be extended into a deep architecture, inspired by recent successes of deep neural networks. We propose a deep boosting framework based on layer-by-layer joint feature boosting and dictionary learning. In each layer, we construct a dictionary of filters by combining the filters from the lower layer, and iteratively optimize the image representation with a joint discriminative-generative formulation, i.e. minimization of empirical classification error plus regularization of analysis image generation over training images. For optimization, we perform two iterating steps: i) to minimize the classification error, select the most discriminative features using the gentle adaboost algorithm; ii) according to the feature selection, update the filters to minimize the regularization on analysis image representation using the gradient descent method. Once the optimization is converged, we learn the higher layer representation in the same way. Our model delivers several distinct advantages. First, our layer-wise optimization provides the potential to build very deep architectures. Second, the generated image representation is compact and meaningful. In several visual recognition tasks, our framework outperforms existing state-of-the-art approaches.
\end{abstract}


\end{frontmatter}


\section{Introduction}\label{sec:Introduction}

Visual recognition is one of the most challenging domains in the field of computer vision and smart computing. Many complex image and video understanding systems employ visual recognition as the basic component for further analysis. Thus the design of robust visual recognition algorithm is becoming a fundamental engineering in computer vision literature and has been attracting many related researchers. Since the inadequate visual representation will greatly influence the performance of visual recognition system, almost all of the related methods are concentrated on developing the effective visual representation.



Traditional visual recognition systems always adopt the shallow model to construct the image/video representation. Among them, the \textit{bag-of-visual-words} (BoW) model, which is the most successful one for visual content representation, has been widely adopted in many computer vision tasks, such as object recognition \cite{PyramidMatch,UniversalVisualDictionary} and image classification \cite{15Scenes,FeiFeiLi}. The basic pipeline of BoW model consists of local feature extraction~\cite{SIFT,HOG}, feature encoding~\cite{ScSPM,LLC,SuperVectorCoding} and pooling operation. In order to improve the performance of BoW, two crucial schemes have been involved. First, the traditional BoW model discards the spatial information of local descriptors, which seriously limited the descriptive power of the feature representation. To overcome this problem, the Spatial Pyramid Matching method was proposed in~\cite{15Scenes} to capture geometrical relationships among local features. Second, dictionaries adopted to encode the local feature in traditional methods are learned in a unsupervised manner and can hardly capture the discriminative visual pattern for each category. This issue inspired a series of works ~\cite{LabelConsistentK-SVD,SuperviseDincionary,FisherDictionary}
to train more discriminative dictionaries via supervised learning, which can be implemented by introducing the discriminative term into dictionary learning phase as the regularization according to various criteria.

 As the research going, the deep models, which can be seen as a type of hierarchical representation \cite{CNN,DBN,CDBN} have played an significant role in computer vision and machine learning literature~\cite{3D-Convolutional,RCNN,ding2015deep} in recent years. Generally, such hierarchical architecture represents different layer of vision primitives such as pixels, edges, object parts and so on~\cite{Visualizing}. The basic principles of such deep models are concentrated on two folds:  (1) layerwise learning philosophy, whose goal is to learn single layer of the model individually and stack them to form the final architecture; (2) feature combination rules, which aim at utilizing the combination (linear or nonlinear) of low layer detected features to construct the high layer impressive features by introducing the activation function.

\begin{figure*}[ht]
\centering
\includegraphics[width=1\textwidth]{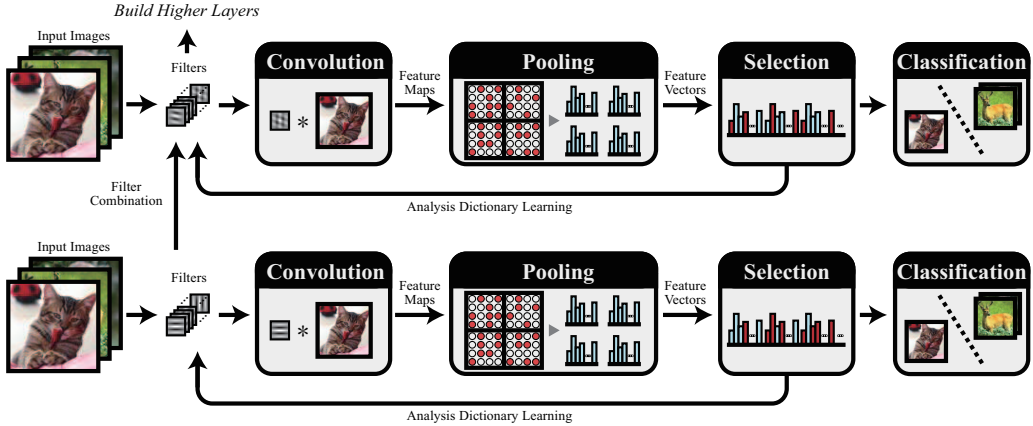}
\caption{A two-layer illustration of proposed deep boosting framework. The horizontal pipelines show the layer-wised image representation via joint feature boosting and analysis dictionary learning. When optimization in the single layer is done, the compositional filters are fed into the higher-layer to generate the novel analysis dictionary for further processing. Note that the feature set in the higher-layer only dependents on the training images and combined filters in the relevant layer. }
\label{pipeline}
\end{figure*}

In this paper, the related exciting researches inspire us to explore how the traditional image classification pipelines, which include feature encoding, spatial pyramid representation and salient pattern extraction (\textit{e.g.}, max spatial pooling operation), can be extended into a deep architecture. To this end, this paper proposes a novel deep boosting framework, which aims to construct the effective discriminative features for image classification task, jointly adopting feature boosting and dictionary learning. For each layer, followed the famous boosting principle~\cite{Gentle-Adaboost}, our proposed method sequentially selects the discriminative visual features to learn the strong classifier by minimizing empirical classification error. On the other hand, the analysis dictionary learning strategy is involved to make the selected features more suitable for the object category. A two-step learning process is investigated to iteratively optimize the objective function. In order to construct high-level discriminative representations, we composite the learned filters corresponding to selected features in the same layer, and feed the compositional results into next layer to build the higher-layer analysis dictionary. Another key to our approach is introducing the model compression strategy when constructing the analysis dictionary, that reduces the complexity of the feature space and shortens the model training time. The experiment shows that our method achieves excellent performance on general object recognition tasks. Fig.~\ref{pipeline} illustrates the pipeline of our deep boosting method (applying two layers as the illustration). Compared with the traditional BoW based method~\cite{ScSPM}, the analysis operation in our model (\textit{i.e.}, convolution) is same as the encoding process that maps the image into the feature space. While the pooling stage is same as the traditional method to compute the histogram representation adopting spatial pyramid matching. Different from traditional models capturing the salient properties of visual patterns by max spatial pooling operation, we adopt the feature boosting to the discriminative features mining for image representation.

The main contributions of this paper are three folds. (1) A novel deep boosting framework is proposed and it leverages the generative and discriminative feature representation. (2) It presents a novel formulation which jointly adopting feature boosting and analysis dictionary learning for image representation. (3) In the experiment on several standard benchmarks, it shows that the learned image representation well discovers the discriminative features and achieves the good performance on various object recognition tasks.

The rest of the paper is organized as follows. Sec.~\ref{sec:RelatedWork} presents a brief review of related work, followed by the overview of background technique details in Sec.~\ref{sec:Background}. Then we introduce our deep boosting framework in Sec.~\ref{sec:Formulation}.
Sec.~\ref{sec:Experiment} gives the experimental results and comparisons. Sec.~\ref{sec:Conclusion} concludes the paper.

\section{Related Work}
\label{sec:RelatedWork}

In the past few decades, many works have been done to design different kinds of features to express the characteristics of the image for further visual tasks. These hand-craft features vary from global expressions~\cite{Feature} to the local representation~\cite{SIFT}. Such designed features can be roughly divided into two types~\cite{HIT}, the one is geometric features and the other is texture features. Geometric features which explicitly record the locations of edges are employed to describe the noticeable structures of local areas. Such features include Canny edge descriptor~\cite{CannyEdge}, Gabor-like primitives~\cite{SparseCoding-Nature} and shape context descriptor~\cite{lin2015discriminatively,luo2015learning}. In contrast, the texture features express the cluttered object appearance by histogram statistics. SIFT~\cite{SIFT}, HoG~\cite{HOG} and GIST~\cite{GIST} are delegates of such feature representation. Beyond such hand-craft feature descriptors, Bag-of-Feature (BoF) model seems to be the most classical image representation method in computer vision area. A lot of illuminating studies \cite{FeiFeiLi,15Scenes,ScSPM,LLC} were published to improve this traditional approach in different aspects. Among these extensions, a class of sparse coding based methods \cite{ScSPM,LLC}, which employ spatial pyramid matching kernel (SPM) proposed by Lazebnik \textit{et al}, has achieved great success in image classification problem. However, despite we are developing more and more effective representation methods, the lack of high-level image expression still plagues us to build up the ideal vision system.


On the other hand, learning hierarchical models to simultaneously construct multiple levels of visual representation has been paid much attention recently. The proposed hierarchical image representation is partially motivated by recent developed deep learning approaches \cite{CNN,DBN,SumProductNetwork}. Different from previous hand-craft feature design method, deep model learns the feature representation from raw data and validly generates the high-level semantic representation. And such abstract semantic representations are expected to provide more intra-class variability. Recently, many vision tasks achieve significant improvement using the convolutional architectures~\cite{3D-Convolutional,RCNN,ding2015deep}. A deep convolutional architecture consists of multiple stacked individual layers, followed by an empirical loss layer. Among all of these layers, the convolutional layer, the feature pooling layer and the full connection layer play major roles in abstract feature representation. The stochastic gradient descent algorithm is always applied to the parameters training in each layers according to back-propagation principle. However, as shown in recent study \cite{SumProductNetwork}, these network-based hierarchical models always contain thousands of parameters. Learning a useful network usually depends on expertise of parameter tuning (\textit{e.g.}, tuning the learning rate and parameter decay rate in each layer ) and is too complex to control in real visual application. In contrast, we build up our hierarchical image representation according to the simple but effective rules. Our method can also achieve the near optimal classification rate in each layer.


Another related work to this paper is learning a dictionary in an analysis prior~\cite{AnalysisSynthesis,SupervisedAnaSyn,AnalysisKSVD}. The key idea of analysis-based model is utilizing analysis operator (also known as analysis dictionary) to deal with latent clean signal and leading to a sparse outcome. In this paper, we consider the analysis-based prior as a regularization prior to learn more discriminative features to a certain category. Please refer to Sec.\ref{sec:Background} for more details about analysis dictionary learning.

\section{Background Overview}
\label{sec:Background}
\subsection{Gentle Adaboost}

We start with a brief review of Gentle Adaboost algorithm \cite{Gentle-Adaboost}. Without loss of generality, considering the two-class classification problem, let $(x_1,y_1)...(x_N,y_N)$ be the training samples, where $x_i$ is a feature representation of the sample and $y_i\in \{-1,1\}$. $w_i$ is the sample weight related to $x_i$. Gentle Adaboost \cite{Gentle-Adaboost,SharingFeatures} provides a simple additive model with the form,
\begin{equation}\label{eq1}
F(x_i) = \sum_{m=1}^{M} f_m(x_i),
\end{equation}
where $f_m$ is called weak classifier in the machine learning literature. It often defines $f_m$  as the regression stump $f_m(x_i)= a\hbar(x_i ^{d}>\delta)+b$, $\hbar(\cdot)$ denotes the indicator function which returns 1 when $x_i ^{d}>\delta$ and 0 otherwise, $x_i ^{d}$ is the $d$-th dimension of the feature vector $x_i$, $\delta$ is a threshold, $a$ and $b$ are two parameters contributing to the linear regression function. In iteration $m$, the algorithm learns the parameter $(d,\delta,a,b)$ of $f_m(\cdot)$ by weighted least-squares of $y_i$ to $x_i$ with weight $w_i$,
\begin{equation}\label{eq2}
\min_{1\leq d \leq D} \sum_{i=1}^N w_i\parallel a^{d}\hbar(x_i^{d} >\delta^{d})+b^{d} - y_i  \parallel ^2 ,
\end{equation}
where $D$ is the dimension of the feature space. In order to give much attention to the cases that are misclassified in each round, Gentle Adaboost adjusts the sample weight in the next iteration as $w_i\leftarrow w_i e^{-y_i f_m(x_i)}$ and updates $F(x_i) \leftarrow F(x_i)+f_m(x_i)$. At last, the algorithm outputs the result of strong classifier as the form of sign function $sign[F(x_i)]$. In this paper,  we adopt Gentle Adaboost as the basic component of proposed model. Please refer to \cite{Gentle-Adaboost,SharingFeatures} for more technique details.

\subsection{Analysis Dictionary Learning}

Our work is also inspired by the recent developed analysis-based sparse representation prior learning~\cite{AnalysisSynthesis,SupervisedAnaSyn,AnalysisKSVD}, which represents the input signal from a dual viewpoint of the commonly used synthesis model~\cite{DictionaryPair}. The main idea of analysis prior leaning is to learn the analysis operators (\textit{e.g.}, convolution operator) that can return the special responses (\textit{e.g.}, sparse response as usual) from the latent signal according to the given constraint. Let $\widehat{I}$ be the observed signal (\textit{e.g.}, natural image) with noisy which is often assumed as zero-mean white Gaussian. An analysis-based prior seeks the latent signal $I$ whose analysis transform result is sparse,

\begin{equation}\label{eq3}
\min_{I,G} \frac{1}{2} \|\widehat{I}-I\|_2^2 + \psi \Phi(G * I ),
\end{equation}
where $\psi\geq0$ is a scalar constant and the symbol $*$ indicates the analysis operation. The first term denotes the reconstruction error and the second one denotes the sparsity constraint of the forward transform coefficient. $G$ is usually a redundant dictionary employing as the analysis operator. In different context, such analysis prior $G$ is more frequently adopted to enforce some regularity on the signal. In this paper, we utilize the philosophy of analysis-based prior to seek the discriminative filters for image feature representation. Please refer to~\cite{AnalysisSynthesis,SupervisedAnaSyn,AnalysisKSVD} for more technique details and theoretical analysis.

\begin{figure*}[!ht]
\centering
\includegraphics[width=4.5in]{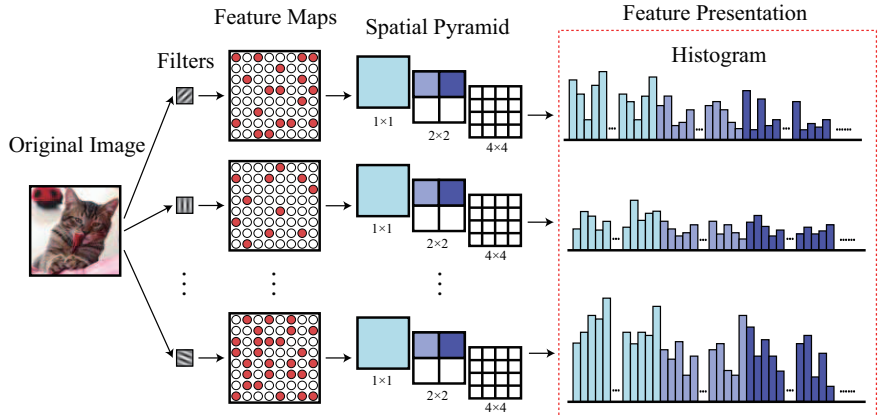}
\caption{Toy example of constructing a three-level pyramid histogram as the image feature representation. }
\label{fig:details}
\end{figure*}

\section{Problem Formulation}
\label{sec:Formulation}

Considering the two-class classification problem, for given training data and its corresponding label $\{(x_i,y_i)|i\in\{1,\ldots,N\}\}, y_i \in \{-1,1\}$. In order to construct the rich and discriminative image representation for each category, we propose a deep boosting framework based on compositional feature selection and analysis dictionary learning. For a single layer, we firstly introduce the term of empirical error to the discriminative features mining. This is equal to learn the weak classifier in Gentle Adaboost algorithm. For each category, suppose that if we can find an analysis dictionary, denoted by $G\in R^{p\times M}$, that the selected feature can be more suitable for such category by the analysis transformation, then the feature representation would be more effective for visual recognition. Based on this idea, the fundamental of our single layer image representation is expressed as follows,

\begin{equation}\label{eq4}
\min_{G}\frac{1}{2} \sum_{i=1}^Nl(-y_iF(x_i))) +\lambda \sum_{I_j\notin \Omega}\| G\ast I_j \|_2^2,
\end{equation}
where $x_i$ is the feature representation corresponding to image $I_i$ and $l(\cdot)$ denotes the empirical error of the classifier. $\Omega$ indicates positive training set and $I_j\notin \Omega$ means that the image $I_j$ does not belong to the set of positive samples. We define $G=[g_1,g_2,...g_m...,g_M]$ as the analysis dictionary and each $g_m$ indicates a linear filter. Thus $G*I$ can be considered as the a series of convolutional operations and the output is $M$ feature maps, each of which is related to a special linear filter. The properties of our proposed model are two folds. On one hand, different from traditional analysis prior learning, we adopt the empirical error, which is more suitable for training the classifier, to replace the reconstruction error in Eq.(\ref{eq2}). On the other hand, the analysis operator is introduced as the regularized term to learn more discriminative features for each category. In the second term of Eq.(\ref{eq4}), we desire the analysis dictionary (\textsl{i.e.}, a set of filters) has large filter response over the positive training set. In this way, the analysis dictionary learning process could discover category coherent features (\textsl{i.e.}, one category one analysis dictionary) to promote the discriminative ability of weak classifiers. It is equivalent to make the analysis dictionary has the small response over negative samples, thus we extract negative training samples and minimize the objective function to train the analysis dictionary. Note that, if the learned filter has the small response to both the positive and negative samples, the related feature representation will be eliminated in the further iteration of feature selection process. In this way, the discriminative of our image representation is enhanced by joint feature boosting and analysis dictionary learning, leading the model more robust and compact as well.

In Eq.(\ref{eq4}), $x_i$ is the feature vector of $i$-th image associated with the analysis transformation (\textit{i.e.}, filter response or convolution result). In order to obtain such feature representation, we employ the pyramid-wise histograms to quantize the filter responses, which provide some degree of translation invariance for the extracted features, as in hand-crafted features (\textit{e.g.}, SIFT or HoG), learned features (\textit{e.g.}, Bag-of-Visual-Words model), and average or maximum pooling process in convolution neural network. Suppose $M$ is the total number of filters. Before construct the pyramid-wise histograms for a special image $I$, we firstly activate the maximum filter responses of each pixel and abandon the others as follows,

\begin{equation}\label{eq5}
u_{m}=
  \left\{
   \begin{aligned}
   \|u_{m}\|   & \;\;\; if \; \|u_{m}\| = \max\{\|u_{1}\|,\|u_{2}\|,...,\|u_{M}\|\}\\
   0         & \;\;\; otherwise\\
   \end{aligned}
   \right.,
  \end{equation}
where $u_m$ indicates the $m$-th filter response of pixel $u \in I$.

According to the previous operation, we can obtain $M$ feature maps for a training image, each of which has only a few locations being activated according to Eq.(\ref{eq5}) (presented by red solid circle in Fig.~\ref{fig:details}). As shown in Fig~.\ref{fig:details}, we apply a three-level spatial pyramid representation of each resulting feature map, resulting $1+2\times2+4\times4=21$ individual spatial blocks. We compute the histogram (with $C$ bins, $C=50$ in the rest of the paper) of the filter responses in each block. Finally, we can get the ``long'' feature vector formed by concatenating the histograms of all blocks from all feature maps. The dimension of such feature vector is $21\times50\times M$. Note that $M$ is not a constant scalar in this paper, and the value could be dynamically changed with the process of analysis dictionary learning. Please refer to Sec.~\ref{sec:DictionaryPursuing} for more details.

\subsection{Feature Boosting}
\label{sec:selection}

In order to optimize the objective function in Eq.(\ref{eq4}), we propose a two-step optimizing strategy integrating the feature boosting and dictionary learning. In this subsection, we describe the details of feature boosting method by setting up the relationship between the weak classifier and the image feature representation. After the pyramid-wise histogram calculated, we select the discriminative features and obtain the single layer classifier through the given feature set. Follow the previous notation, let $x_i \in R^D$ be the feature representation of image $I_i$, where $D$ is the dimension of the feature space and $D=21\times50\times M$ as described in the previous content. In the feature boosting phase, Gentle Adaboost is applied to the discriminative features (\textit{i.e.}, weak classifiers ) mining, which can separate the positive and negative samples nicely in each round. Note that in the rest of the paper, we apply $x_i^d$ to denote the value of $x_i$ in the $d$-th dimension. In each round of feature boosting procedure, the algorithm retrieves all of the candidate regression functions $\{f^1,f^2,...,f^D\}$, each of which is formulated as:

\begin{equation}\label{eq6}
f^d(x_i)= a\phi(x_i^{d}-\delta)+b,
\end{equation}
where $\phi(\cdot)$ is the sigmoid function with the form $\phi(x)=1/(1+e^{-x})$. For each round, the candidate function with minimum empirical error is selected as the current weak classifier $f$, such that

\begin{equation}\label{eq7}
\min_{d} \sum_{i=1}^N w_i\parallel f^d(x_i) - y_i  \parallel ^2 ,
\end{equation}
where $f^d(x_i)$ is associated with the $d$-th element of $x_i$ and the function parameter $(\delta,a,b)$. According to the above discussion, we build the bridge between the weak classifier and the feature representation, thus the weak classifiers learning can be viewed as the feature boosting procedure in our model. The feature boosting is usually terminated when the training error is converged.

\subsection{Analysis Dictionary Learning}
\label{sec:DictionaryPursuing}

To the regularization perspective, another advantage of method is introducing analysis dictionary learning, which is conducted by selected features in the feature boosting phase, to emphasize the discriminative ability of analysis operator for the target category. In our framework, since we rely on discriminative filters to generate higher-layer proper analysis dictionary, we only consider to update a subset of filters which is corresponding to the selected features. We first need to construct the relationship between feature responses and filters. For any feature response, a four-item index is recorded as,

\begin{equation}\label{eq8}
[isActivited, w,h,g] ,
\end{equation}
where $isActivited$ indicates whether the feature response is selected in feature boosting stage. $w,h$ are the horizontal and vertical coordinate in the image lattice domain respectively. $g$ denotes the relative filter defined in Eq.(\ref{eq4}). Then we apply the gradient descent algorithm to optimize filters which is corresponding to selected features. As Fig.\ref{pipeline} illustrates, we combine any two optimized filters but not the features to generate filters in the next layer. In this way, the filter's optimization in the next layer is independent with previous features. Note that in the first few layers, the number of filters is limited, thus almost every filter is taken into account in optimization. However, it will show in Sec.\ref{sec:composite} that the collection of compositional filters becomes large along with the architecture going deep, thus the screening mechanism is introduced to control the complexity and keep the effectiveness of the model.

Integrating the two stages described in Sec.~\ref{sec:selection} and Sec.~\ref{sec:DictionaryPursuing}, we achieve the feature boosting and analysis dictionary learning for the single layer. The algorithm is summarized in Alg.\ref{alg:SingleLayer}. In next subsection we will introduce the filter combination rules to construct the hierarchical architecture of our model.

\subsection{Deep Boosting Framework}
\label{sec:composite}

\begin{figure}[ht]
\centering
\subfigure[Illustration of compositional filters.]{
\begin{minipage}[b]{0.7\textwidth}
\centering
\includegraphics[width=1\textwidth]{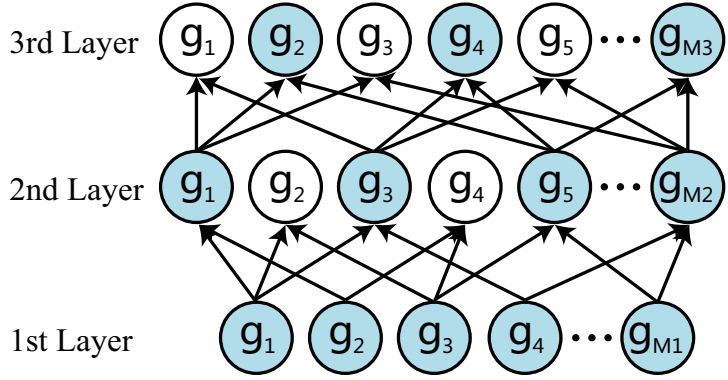}
\end{minipage}
\label{filterCom-a}
}

\subfigure[The similarity matrix of \nth{2} layer.]{
\begin{minipage}[b]{0.3\textwidth}
\centering
\includegraphics[width=1\textwidth]{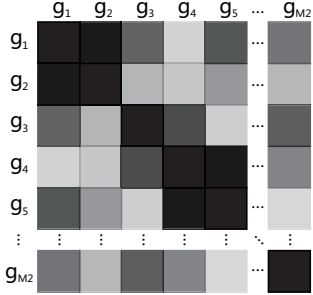}
\end{minipage}
\label{filterCom-b}
}
\hspace{0.4in}
\subfigure[The similarity matrix of \nth{3} layer.]{
\begin{minipage}[b]{0.3\textwidth}
\centering
\includegraphics[width=1\textwidth]{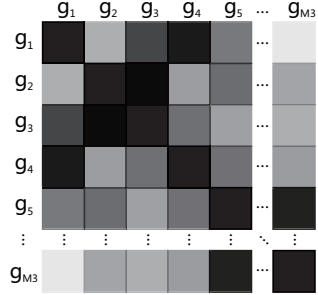}
\end{minipage}
\label{filterCom-c}
}
\caption{Illustration of compositional filters for deep boosting. We composite filters in a pairwise manners in each layer and treat the output compositional filters as base filters (presented by solid circle in Fig.~\ref{filterCom-a}) in next layer. After combination, the similar matrix of filters is built up to drop out redundancies (presented by hollow circle in Fig.~\ref{filterCom-a}).}
\label{filterCom}
\end{figure}

In the context of boosting method, the strong classifier, which is usually the weighted linear combination of weak classifiers, is hardly to decease the test error when training error is approaching to zero. Based on this fact, it is our interest to learn high-level feature representations with more discriminative ability. In order to achieve this goal, we propose the filter combination rules and the output compositional filters of each layer are treated as a whole to generate the analysis dictionary in the next layer.

For each image category, whose corresponding analysis dictionary in layer $l$ is denoted by $[G]_{l}$, we combine any two optimized filters (presented by solid circle in Fig.~\ref{filterCom-a}) in the $l$-th layer as follows,

\begin{equation}\label{eq9}
[g_k]_{l+1} = \phi( \: [g_i]_{l} + [g_j]_{l}\:),
\end{equation}
where $\phi(\cdot)$ is the sigmoid function. $[g_i]_{l}$ and $[g_j]_{l}$ indicate the $i$-th and $j$-th filters in the optimized subset of $[G]_{l}$. As illustrated in Fig.~\ref{filterCom-a}, the number of filters in each layer is quite different and we only adopt the optimized ones, which are related to selected features, to construct the image filters for the next layer.

\subsection{Model Compression Approach}

Although we carefully select filters for further combination, the number of compositional filters will still be out of control when architecture going deep. Assuming there exists $M_l$ optimized filters in layer $l$, thus we can obtain he maximum number $\frac{1}{2}\times M_l\times(M_l-1)$ of compositional filters. In this way, the dimension of each image in the layer $l+1$ would be $ \frac{1}{2}\times M_l\times(M_l-1)\times 21 \times 50 $, which make the feature space is too complex and the training time becomes intolerable. To this end, we introduce model compression in the training phase. For any couple of filers, the L2 distance is calculated to measure the similarity between them. If the distance is smaller than the threshold $\delta$ (set as $0.7$ in all the experiment), we maintain the two filters are similar and one of them is dropped out randomly (presented by hollow circle in Fig.~\ref{filterCom-a}). Fig.~\ref{filterCom-b} and Fig.~\ref{filterCom-c} illustrate the similarity matrix of filters in different layer. The intensity of every square indicates the similar degree of two filters. Please refer to Fig.~\ref{fig:modelcomp-acc} and Fig.~\ref{fig:modelcomp-time} for more details about the classification accuracy and training time comparison with and without model compression for different depth of proposed framework.

 According to Sec.~\ref{sec:composite}, we build up the hierarchical architecture of our deep boosting framework. In the testing phase, we employ the weak classifiers learned in every layer to produce the final classifier. The overall of our proposed method is summarized in Alg.~\ref{alg:Framwork}.

\begin{algorithm}[!ht]
\caption{Joint Feature Boosting and Analysis Dictionary Learning}
\label{alg:SingleLayer}
\begin{algorithmic}
\REQUIRE ~~\\
    Positive and negative training samples $(x_1,y_1)...(x_N,y_N)$, the number of selected features $\Pi$.
\ENSURE ~~\\                           
    A pool of selected features $\Psi$, the learned dictionary $G$.
\INPUT ~~\\
    The dictionary $G$;
\MYWHILE
    \STATE
    \begin{itemize}
    \setlength{\itemsep}{1pt}
    \setlength{\parskip}{3pt}
    \setlength{\parsep}{10pt}
    \item[1.] Start with score $F(x)=0$ and sample weights $w_i=1/N$, $i=1,2,\ldots,N$.
    \item[2.] Select features and learn the strong classifier as follows:

        \textbf{Repeat} for $m=1,2,\ldots,\Pi$:
        \setlength{\parskip}{-1pt}
            \begin{itemize}
            \setlength{\itemsep}{1pt}
            \setlength{\parskip}{0pt}
            \setlength{\parsep}{10pt}
            \item[(a)] Learn the current weak classifier $f_m$ by Eq.(\ref{eq6}).
            \item[(b)] Update $w_i\leftarrow w_i e^{-y_i f_m(x)}$ and renormalize.
            \item[(c)] Update $F(x) \leftarrow F(x)+f_m(x)$.
            \end{itemize}
        \setlength{\parskip}{0pt}

    \item[3.] Update the dictionary $G$ by gradient descent method.
    \item[4.] Generate new feature vectors of each image using $G$ according to Sec.~\ref{sec:Formulation}.
    \end{itemize}

    \MYENDWHILE

 The objective function in Eq.(\ref{eq4}) converges.

\end{algorithmic}
\end{algorithm}

\begin{algorithm}[!ht]
\caption{Deep Boosting Framework}
\label{alg:Framwork}
\begin{algorithmic}
\REQUIRE ~~\\
    Positive and negative training images and corresponding labels $(I_1,y_1)...(I_N,y_N)$, the number of selected features $\Pi_l$ in layer $l$, the total layer number $L$.
\ENSURE ~~\\                           
    The final classifier $F^L(x)$ for a special category.\\
\INPUT ~~\\
    Initialize $G^{'}$ in first layer applying Gabor wavelets.
\MYWHILE for $l=1,2,\ldots,L$:
    \STATE
    \begin{itemize}
        \item[1.] Generate new feature $x$ of image $I$ using $G$ according to Sec.~\ref{sec:Formulation}.
        \item[2.] Boost features with dictionary learning according to Alg.~\ref{alg:SingleLayer}.
        \item[3.] Build up filters of next layers according to Eq.(\ref{eq9}).
    \end{itemize}

\end{algorithmic}
\end{algorithm}

\subsection{Preprocessing and Multi-class Decision}

At the beginning, we initialize the filters with the size of $5\times 5$ adopting Gabor wavelets. Let $I$ be an image defined on image lattice domain and $G^{'}$ be the Gabor wavelet elements with parameters $(w,h,\alpha,s)$, where $(w,h)$ is the central position belonging to the lattice domain, $\alpha$ and $s$ denote the orientation and scale parameters. Different orientation and scale parameters makes Gabor wavelets variant. For simplicity, we apply 1 scale and 16 orientations in our implementation, so there are total 16 filters at first layer. Notably, multi scales promote the performance while the filter combination process becomes complicated, because the combination is only allowed in the same scale. Followed by \cite{ActiveBasisModel}, we utilize the normalize term to make the Gabor responses comparable during the inception phase between different training images:

\begin{equation}\label{eq10}
\delta^2(s) = \frac{1}{|P|A}\sum_{\alpha}\sum_{w,h} |\langle I,G^{'}_{w,h,\alpha,s}\rangle|^2 ,
\end{equation}
where $|P|$ is the total number of pixels in image $I$, and $A$ is the number of orientations. $\langle\cdot\rangle$ denotes the convolution process. For each image $I$, we normalize the local energy as $|\langle I,G^{'}_{w,h,\alpha,s}\rangle|^2 / \delta^2(s)$ and define positive square root of such normalized result as feature response.

\begin{figure}[!ht]
\centering
\includegraphics[width=0.65\textwidth]{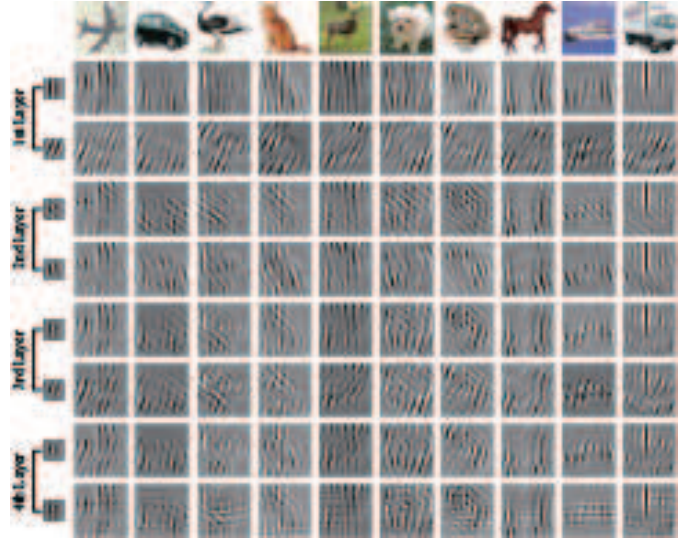}
\caption{The learned templates in the first four layers for each image categories. When the model goes deeper, we get higher level primitives and the more discriminative features.}
\label{fig:template}
\end{figure}

To the multiclass situation, we consider the naive \textit{one-vs-all} scheme to train multiple binary classifiers, each one learns to distinguish the samples in a single class from the samples in all remaining classes. Given the training data $\{(x_i,y_i)\}_{i=1}^N$,$y_i\in \{1,2,...,K\}$, we train $K$ strong classifiers, each of which returns a classification score for a special test image. In the testing phase, we predict the label of image referring to the classifier with the maximum score. The reason why we adopt \textit{one-vs-all} or OVA scheme throughout the paper is concentrated on two folds. On one hand, according to the Eq.(\ref{eq4}), we desire each learned analysis dictionary should have powerful capability to distinguish the images from one category. Thus we select the negative samples from all other categories to optimize the filters in Eq.(\ref{eq4}) (\textit{i.e.}, leaning the class-specific analysis dictionary) and this strategy is naturally consistent with the OVA scheme. On the other hand, as shown in \cite{DBLP:DefenseOfOne-Vs-All}, many multiclass models may not offer advantages over the simple OVA scheme in the solution of classification problem. Under such circumstances, we finally choose the OVA strategy followed by its intuitive concept.

\section{Experiment}\label{sec:Experiment}

We conduct several experiments to investigate the properties of proposed deep boosting framework and evaluate the performance for different challenging visual recognition tasks (\textit{i.e.}, facial age estimation, natural image classification and similar appearance categories recognition). All of the experiments are carried out on a PC with Core i7-3960X 3.30 GHZ CPU and 24GB memory. In these tasks, we demonstrate superior or comparable performances of our framework over other state-of-the-art approaches.

\subsection{Learning image template for image categories}

In the first experiment we focus on whether our algorithm can learn and select meaningful and discriminative features for different image categories. Take CIFAR-10 dataset, for example. The CIFAR-10 dataset \footnote{http://www.cs.toronto.edu/$\sim$kriz/cifar.html} consists of 60K $32\times32$ color images in 10 classes (with 6K images per class), including airplane, automobile, bird, cat, deer, dog, frog, horse, ship and truck. We randomly select 1,000 images per class as the training samples to learn the hierarchical image representation. Fig.\ref{fig:template} shows some learned templates in different layers for each image categories. According to the visualizations, it is obviously that the higher layer it goes, the more informative features we gain.

\begin{figure}[!ht]
\centering
\includegraphics[width=0.5\textwidth]{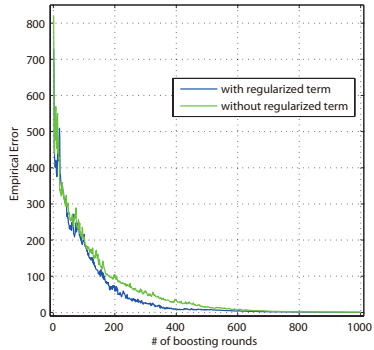}
\caption{The empirical error at boosting rounds. The method with regularized term has better convergence rate.}
\label{fig:EEvsBR}
\end{figure}

\begin{figure}[!ht]
\centering
\includegraphics[width=0.5\textwidth]{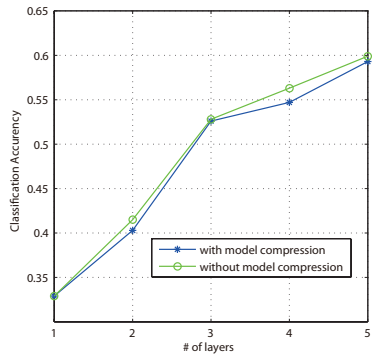}
\caption{Classification accuracy at different layers. The method conduct better performance with the growth of model.}
\label{fig:modelcomp-acc}
\end{figure}

\subsection{Natural image classification}
The same to CIFAR-10, the STL-10\footnote{http://cs.stanford.edu/$\sim$acoates/stl10/} is also a ten-category image dataset, but with the image size $96\times96$. It has 1300 images per class. There are 500 training images and 800 test images. The training set is mapped to ten predefined folds. Due to its relatively large image size, much prior research chose to downsample the images to $32\times32$. Tab.~\ref{table:stl10} shows the comparison of average test accuracies on all folds of STL-10. It is clear that our method can achieve very competitive results compared to other state-of-the-art methods.

\begin{table}[ht]
\caption{Classification accuracy on STL-10.}
\centering
\begin{tabular}{|c|c|}
\hline
\textbf{Method} & \textbf{Accuracy ($\pm  \sigma$)} \\
\hline
1-layer Vector Quantization\cite{SCVQ}& 54.9\% ($\pm$ 0.4\%) \\
1-layer Sparse Coding\cite{SCVQ}  & 59.0\% ($\pm$ 0.8\%) \\
3-layer Learned Receptive Field\cite{ReceptiveFields}  & 60.1\% ($\pm$ 1.0\%)  \\
OURS-5  & 59.3\% ($\pm$ 0.8\%)  \\
\hline
\end{tabular}
\label{table:stl10}
\end{table}

\begin{figure}[!ht]
\centering
\includegraphics[width=0.5\textwidth]{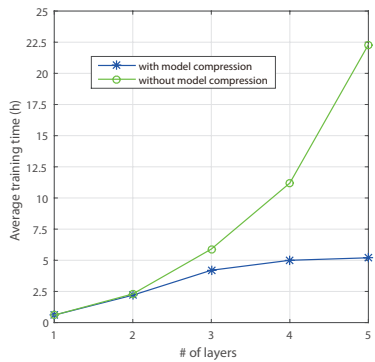}
\caption{The average training time of categories at different layers. The average training time of categories greatly reduce when the model is compressed.}
\label{fig:modelcomp-time}
\end{figure}

\begin{figure*}[!ht]
\centering
\includegraphics[width=1\textwidth]{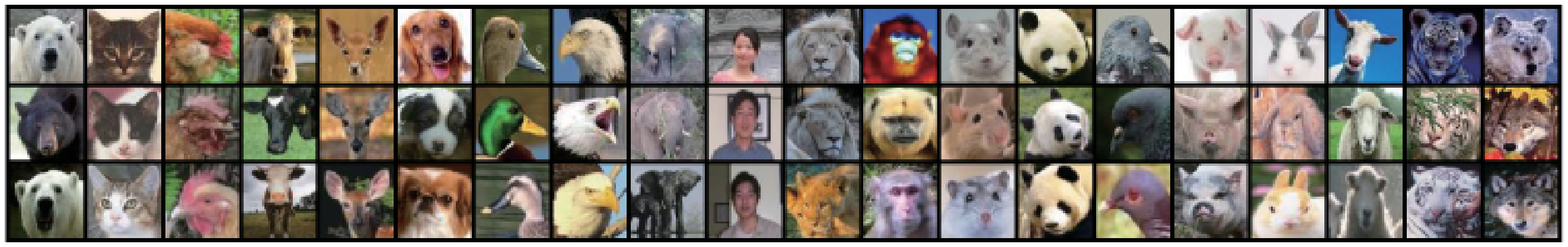}
\caption{The LHI-Animal-Faces dataset. Three images are shown for each category. }
\label{fig:animalface}
\end{figure*}

\begin{figure}[ht]
\centering
\subfigure[The original images.]{
\begin{minipage}[b]{0.6\textwidth}
\centering
\includegraphics[width=1\textwidth]{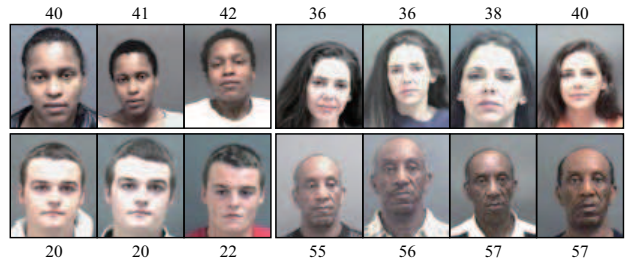}
\end{minipage}
\label{morph-a}
}
\subfigure[The aligned and cropped images.]{
\begin{minipage}[b]{0.6\textwidth}
\centering
\includegraphics[width=1\textwidth]{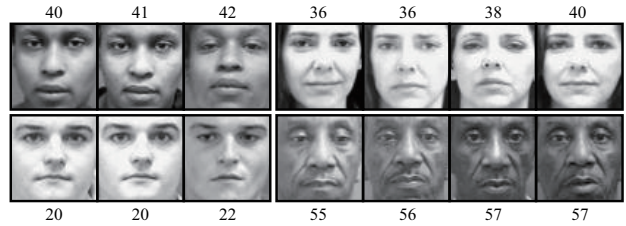}
\end{minipage}
\label{morph-b}
}
\caption{The MORPH-\uppercase\expandafter{\romannumeral2} dataset. Four individuals in different races and genders are picked as an example. The ages are given around the images.}
\label{morph}
\end{figure}

\begin{figure}[!ht]
\centering
\includegraphics[width=0.5\textwidth]{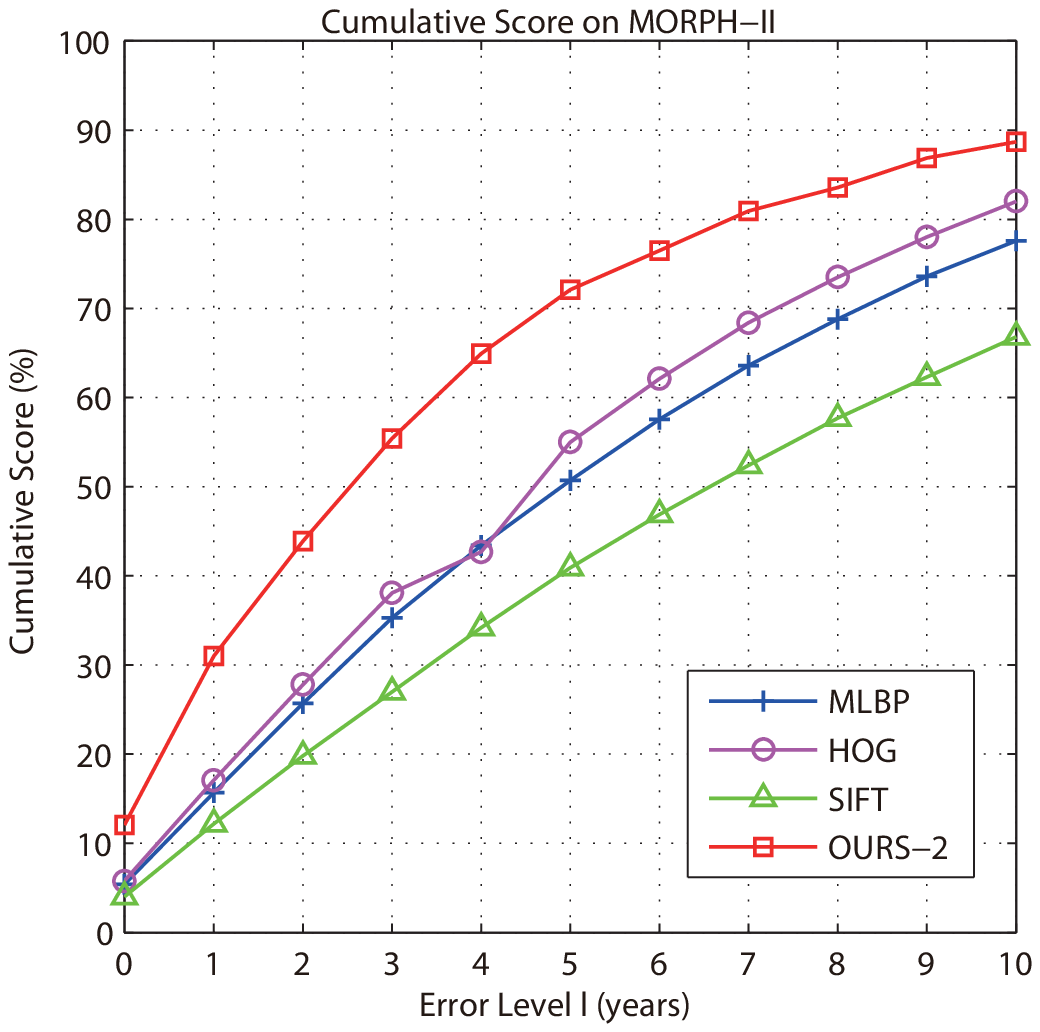}
\caption{Cumulative scores  at different error levels on MORPH-\uppercase\expandafter{\romannumeral2}. }
\label{fig:morphCS}
\end{figure}

\subsubsection{Impact of analysis dictionary learning}

In this section, we are interested in the performance of our method in context of analysis dictionary learning. As we mentioned above, the analysis operator is introduced as a regularized term to learn more discriminative features over the positive samples. We desire that the analysis dictionary is able to make the margin between positive and negative training sets as larger as possible. That is, the analysis dictionary has large response over the positive training set, but not vice versa. Note that, the related feature representation will be eliminated in the further iteration of feature selection process, if the learned filter responds a small value both to the negative set and to the positive set. In this way, we will gain more discriminative features in feature boosting procedure, resulting a more robust and compact image representation model.

Tab.~\ref{table:stl10rt} shows the classification accuracy with and without regularized term. The result using regularized term outperforms the other and the standard deviation among folds is smaller, which illustrates that the feature is more discriminative and the model is more robust. In Fig.~\ref{fig:EEvsBR}, the empirical error in boosting phase is shown. For the more discriminative features, it is reasonable to accelerate convergence rate using regularized term.

\begin{table}[ht]
\caption{Classification accuracy on STL-10 dataset with and without regularized term. }
\centering
\begin{tabular}{|c|c|}
\hline
 &  Accuracy ($\pm \sigma$)\\
\hline
with regularized term &  59.3\% ($\pm$  0.8\%)  \\
without regularized term &  55.8\% ($\pm$  1.5\%) \\
\hline
\end{tabular}
\label{table:stl10rt}
\end{table}

\subsubsection{Impact of model depth and compression}

In this experiment, we perform classification experiments on the STL-10 in the context of different number of layers. We learn the deep boosting model to construct multiple levels of visual representation simultaneously. In order to construct high-level discriminative representations, we composite the learned filters corresponding to selected features in the same layer, and feed the compositional results into next layer to build the higher-layer analysis dictionary. Hopefully when the model goes higher, the features is more discriminative. Fig.~\ref{fig:modelcomp-acc} exhibits the performance of image classification on STL-10 at different layers. The results demonstrate that the features in higher layer conduct better performance. In order to avoid the sudden explosion of
filters, we drop out similar filters randomly after pairwise combination of the learned filters. Although it losses accuracy slightly, we control the training time and make the limitless growth of model possible, which is illustrated in Fig.~\ref{fig:modelcomp-acc} and Fig.~\ref{fig:modelcomp-time}.

\subsection{Similar appearance categories recognition}

The LHI-Animal-Faces dataset\footnote{http://www.stat.ucla.edu/$\sim$zzsi/hit/changelog.html} \cite{HIT} consists of about 2200 images for 20 categories. Fig.~\ref{fig:animalface} provides an overview of the dataset. In contrast with other general classification datasets, LHI-Animal-Faces contains only animal or human faces, which are similar to each other. It is challenging to discern them for their evolutional relationship and shared parts. Besides, interesting within-class variation is shown in the face categories, including rotation, flip transforms, posture variation and sub-types.

We compare our result with those reported in \cite{AOT} obtained by other methods, which include HoG feature trained with SVM \cite{HOG}, HIT \cite{HIT}, AOT \cite{AOT} and partbased HoG feature trained with latent SVM \cite{partBasedHoG}. In experiment, we splits the dataset as training set and test set following AOT \cite{AOT}. For our method, we resize all the images to the uniform size of $60\times 60$ pixels and the number of layers is $5$. Tab.~\ref{table:animalfaces} exhibits the classification accurracy on LHI-Animal-Faces. It has shown that our method achieves a $2.4\%$ increase, compared with the second best competitor.

\begin{table}[ht]
\caption{Classification accuracy on LHI-Animal-Faces. }
\centering
\begin{tabular}{|c|c|}
\hline
\textbf{Method} & \textbf{Accuracy} \\
\hline
HoG+SVM & 70.8\% \\
HIT\cite{HIT} &  75.6\% \\
LSVM\cite{partBasedHoG} &  77.6\% \\
AOT\cite{AOT} &  79.1\% \\
\textbf{OURS-5} &  \textbf{81.5\%} \\
\hline
\end{tabular}
\label{table:animalfaces}
\end{table}

\subsection{Facial age estimation}

Human age estimation based on facial images plays an important role in many applications, \textit{e.g.}, intelligent advertisement, security surveillance monitoring and automatic face simulation. To our best knowledge, MORPH-\uppercase\expandafter{\romannumeral2}\footnote{http://www.faceaginggroup.com/morph/} is the largest publicly available dataset for facial age estimation. In the MORPH-\uppercase\expandafter{\romannumeral2} dataset, there are more than $55,000$ facial images from more than $13,000$ individuals with only about 4 labeled images per individual. The ages vary over a wide range from 16 to 77. The individuals come from different races, among them Africans accounted for about $77\%$, the Europeans about $19\%$, and the remaining includes
Hispanic, Asian and other races. Some sample images are shown in Fig.~\ref{morph-a}.

We use two usually performance measures in our comparative study, \textit{i.e.}, MAE (Mean Absolute Error) and CumScore (Cumulative Score)~\cite{AgeEsti}. Suppose there are $N$ test images, the MAE is the sum of average absolute errors between the true ages $a_i$ and the predicted ages $\bar{a}_i$, $i=1, 2, \cdots, N$. The MAE is calculated as,

\begin{equation}
MAE=\frac{1}{N}\sum_i^N |a_i-\bar{a}_i|,
\end{equation}
where $|\cdot|$ denotes the absolute value of a scalar value.

The CumScore is the cumulate accuracy rate. A certain error range (\textit{i.e.}, $l$ years) is acceptable for many real applications. The cumulative score at error level $l$ can be calculated as,

\begin{equation}\label{eq11}
CumScore(l) = N_{e\leqslant l} / N \times 100\%,
\end{equation}
where $N_{e\leqslant l}$ is the number of test images, which have absolute prediction error no more than $l$ years.

For an input image, we locate the face with bounding box and detect the five facial key points in the bounding box. The five facial key points include two eye centers, nose tip, and two mouth corners. Then we align the facial image based on these key points. Finally, the images are resized to the size of $60\times60$ pixels. The aligned images are shown in Fig.~\ref{morph-b}.

We compare our results with several existing algorithms designed for the age estimation, \textit{i.e.}, IIS-LLD \cite{IIS-LLD}, WAS \cite{WAS} and AGES \cite{AGES}. Moreover, we also conduct experiments using some feature descriptors usually used in face recognition, including Multi-level LBP \cite{Multi-levelLBP}, HoG \cite{HOG} and SIFT \cite{SIFT}. For all of these features, age estimation is treated as classification problem using multi-class SVMs. For our method, we set the number of layers to 2 and six-folder cross validation is performed.  Tab.~\ref{table:morph} summarizes the results based on the MAE measure. We can see that our method achieves better results compared to other state-of-the-art methods for age estimation. We also report the results in terms of the cumulative scores at different error levels from $0$ to $10$ in Fig.~\ref{fig:morphCS}, exhibiting that our method outperforms other state-of-the-arts at almost all levels.

\begin{table}[ht]
\caption{MAE (in Years) on MORPH-\uppercase\expandafter{\romannumeral2} (the lower the better).}
\centering
\begin{tabular}{|c|c|}
\hline
\textbf{Method} & \textbf{MAE} \\
\hline
MLBP+SVM & 6.85 \\
HoG+SVM &  6.19 \\
SIFT+SVM &  8.77 \\
WAS\cite{WAS} & 9.21 \\
AGES\cite{AGES} &  6.61 \\
IIS-LLD\cite{IIS-LLD} & 5.67 \\
OURS-2 &  \textbf{5.61} \\
\hline
\end{tabular}
\label{table:morph}
\end{table}

\section{Conclusion}
\label{sec:Conclusion}

In this paper, we propose a novel deep boosting framework, which is applied to construct the high-level discriminative features for general image recognition task. For each layer, the feature boosting and analysis dictionary learning are integrated into a unified framework for discriminative feature selection and learning. In order to construct high-level image representation, the combined filters in the same layer are fed into next layer to generate the novel analysis dictionary. The experiments in several benchmarks demonstrate the effectiveness of proposed method and achieve good performance on various visual recognition tasks.

\section*{Acknowledgement}
This work was supported by the National Natural Science Foundation of China (no. 61170193, no. 61370185), Guangdong Science and Technology Program (no. 2012B031500006), Guangdong Natural Science Foundation (no. S2012020011081, no. S2013010013432), Special Project on Integration of Industry, Education and Research of Guangdong Province (No. 2012B091000101), and Program of Guangzhou Zhujiang Star of Science and Technology (No. 2013J2200067). Corresponding authors of this work is Liang Lin.

\section*{References}

\bibliography{mybibfile}

\end{document}